\documentclass{article}
\usepackage{spconf,amsmath,graphicx}
\usepackage{microtype}
\usepackage{graphicx}
\usepackage{subfigure}
\usepackage{booktabs} 
\usepackage{hyperref}
\usepackage{amsmath}
\usepackage{amssymb}
\usepackage{mathtools}
\usepackage{amsthm}
\usepackage[capitalize,noabbrev]{cleveref}

\usepackage{array}
\usepackage{threeparttable}
\usepackage{lipsum}
\usepackage{arydshln}

\usepackage{amsfonts,amssymb}
\usepackage{multirow}
\usepackage{pifont}
\usepackage{booktabs}

\newcolumntype{C}{>{$\displaystyle}c<{$}}

\usepackage{makecell}
\usepackage{dsfont}

\title{Understanding Data Augmentation from a Robustness Perspective}

\vspace{-4mm}
\name{Zhendong Liu, Jie Zhang, Qiangqiang He, Chongjun Wang $^{\dagger}$ \thanks{This paper is supported by the National Natural Science Foundation of China (Grant No. 62192783, 62376117), the Collaborative Innovation Center of Novel Software Technology and Industrialization at Nanjing University.}}

\address{Nanjing University\\
Department of Computer Science and Technology\\
Nanjing, China}
\begin{document}
\ninept
\maketitle
\vspace{-4mm}
\begin{abstract}
In the realm of visual recognition, data augmentation stands out as a pivotal technique to amplify model robustness. Yet, a considerable number of existing methodologies lean heavily on heuristic foundations, rendering their intrinsic mechanisms ambiguous. This manuscript takes both a theoretical and empirical approach to understanding the phenomenon. Theoretically, we frame the discourse around data augmentation within game theory's constructs. Venturing deeper, our empirical evaluations dissect the intricate mechanisms of emblematic data augmentation strategies, illuminating that these techniques primarily stimulate mid- and high-order game interactions. Beyond the foundational exploration, our experiments span multiple datasets and diverse augmentation techniques, underscoring the universal applicability of our findings. Recognizing the vast array of robustness metrics with intricate correlations, we unveil a streamlined proxy. This proxy not only simplifies robustness assessment but also offers invaluable insights, shedding light on the inherent dynamics of model game interactions and their relation to overarching system robustness. These insights provide a novel lens through which we can re-evaluate model safety and robustness in visual recognition tasks.
\end{abstract}
\begin{keywords}
Deep learning, Game theory, Data augmentation, Explainability, Robustness
\end{keywords}
\vspace{-2mm}
\section{Introduction}
\vspace{-2mm}
\label{sec:intro}

Deep Neural Networks (DNNs), with their remarkable success in visual recognition, have witnessed increasing integration into real-world applications. Such integration, however, has exposed challenges in model robustness, distribution shifts, and adversarial defense. A salient method to address these challenges is data augmentation, which has proved invisible across various AI architectures, including not only convolutional neural networks (CNNs) but also contemporary vision transformer models \cite{mao2022enhance,zhou2022understanding}.

Recent studies have highlighted the efficacy of data augmentation \cite{hendrycks2021many,hendrycks2018benchmarking,hendrycks2022pixmix}. \cite{yin2019fourier} underscored its role in bolstering robustness against corrupted data. Techniques ranging from basic image transformations to advanced methods like Cutout \cite{devries2017improved}, Pixmix \cite{hendrycks2022pixmix} and Mixup \cite{zhang2018mixup} have been developed based on intuitive principles. Adversarial training, a specialized form of data augmentation, has been investigated using game interactions \cite{ren2021unified}. Although some researchers have tried to summarize data augmentation methods \cite{chen2020group,yang2022sample,mintun2021interaction,dao2019kernel}, a consolidated understanding of the underlying robustness mechanisms is still lacking. 

Game theory has been widely applied to the study of properties of DNNs. A very popular method based on Shapley values \cite{kuhn1953contributions} is SHAP \cite{lundberg2017unified} which unifies six existing XAI methods. There are some works that explain some artificial intelligence techniques from the perspective of game interaction. \cite{zhang2020interpreting} propose the multi-order interaction and try to understand and improve the dropout operation. \cite{ren2021unified} use the multi-order interaction to study the adversarial robustness of DNNs. \cite{deng2021discovering} use the multi-order interaction to study the representation bottleneck of DNNs and observe a phenomenon that indicates a cognition gap between DNNs and human beings.  With a focus on understanding and bridging the cognitive gap between DNNs and humans, we delve into the multi-order game interactions to explore the mechanism of data augmentation.
 
Our contributions include:
\vspace{-2mm}
\begin{itemize}
\item Analysis of popular data augmentation techniques through game interaction theory.
\vspace{-1mm}
\item Introduction of a unified robustness proxy, paving the way for future AI safety research.
\vspace{-1mm}
\item Empirical validation of our theoretical part and proxy.
\vspace{-2mm}
\end{itemize}
\vspace{-2mm}
\section{Theoretical Part}
\vspace{-2mm}
\label{app_a}
 We analyze several existing data augmentation methods and illustrate what these several data augmentation methods have in common. First, we provide a background description of the concept of game interaction to facilitate the reader's understanding. From a game-theoretic perspective, in a set $N=\{1,2,...,n\}$ with $n$ players, we denote all possible subsets of $N$ by $S, S \subseteq N$. Then we switch perspective and consider $n$ players as $n$ input variables of DNN. Among these input variables, two input variables often influence each other. For example, when we train a classifier to recognize whether a face wears lipstick, the eyes are often associated with the mouth, because people with eye makeup generally wear lipstick. In this case, the two input variables can be viewed as a single player $S_{\{ij\}} = \{i, j\}$. Thus, the interaction of the two input variables is defined by \cite{grabisch1999axiomatic}  as follows. 
\begin{align}
       I(i,j) \overset{def}{=} \phi (S_{ij}|N') - [\phi(i|N \backslash \{j\}) + \phi(j|N \backslash \{i\})]  \\
    =  \sum_{S \subseteq N \backslash \{ i,j \}} P_{Shapley}(S|N \backslash \{ i,j \} ) \Delta f(S,i,j)
\end{align}
where $\Delta f(S,i,j) \overset{def}{=} f(S \cup \{i,j \}) - f(S \cup {j}) - f(S \cup {i}) + f(S)$. $\phi(j|N \backslash \{i\})$ and $\phi(i|N \backslash \{j\})$ are the contribution to the DNN output when i and j work individually. $I(i,j)$ is also equal to the change of the variable $i$’s Shapley value when we mask another input
variable $j$ $w.r.t.$ the case when we do not mask $j$.
We use the absolute value of the interaction, $i.e. |I(i,j)|$ to compute the strength of the interaction. 

The multi-order interaction $I^{(m)}(i,j)$ between two input variables $i,j \in N $ proposed by \cite{zhang2020interpreting} is defined as follows:
\begin{equation}
\label{eq1}
    I^{(m)}(i,j) = \mathbb{E}_{S\subseteq N \backslash \{i,j\}, |S|=m}
    [\Delta v(i,j,S)],\ 0 \leq m \leq n-2.
\end{equation}
where $\Delta v(i,j,S) = v(S\cup \{i,j\})  - v(S \cup {i}) 
  - v(S \cup {j}) + v(S)$. The $v(S)$ is the score that a DNN outputs when using variables in the subset $S\subseteq N$ only. The network outputs of a DNN can be explained as follows:
\begin{equation}
\label{eq2}
    v(N)=v(\emptyset) + \sum_{i \in N} \mu_i
    + \sum_{i,j \in N,i \neq j} \sum_{m=0}^{n-2}
    w^{(m)}I^{(m)}(i,j)
\end{equation}
where $\mu_i = v({i})- v(\emptyset)$, $ w^{(m)} = \frac{n-1-m}{n(n-1)}$.
\textbf{When $m$ is small, for example, $m=0.1n$, we call the influence of $m$-order game interactions as low-order interactions.} Similarly, we can define mid-order and high-order interactions. \textbf{For example, $m \in (0.3n, 0.5n)$, we call the influence of $m$-order game interactions as mid-order interactions.} 
  
 How to measure the interaction of different orders interactions is also solved in \cite{deng2021discovering}, as follows.
\begin{align}
       J^{(m)} = \frac{\mathbb{E}_{x \in \Omega} [ 
                       \mathbb{E}_{i,j} [|I^{(m)}(i,j|x)|]
                   ]}
                    {
                    \mathbb{E}_{m'}[ \mathbb{E}_{x \in \Omega} [ 
                        \mathbb{E}_{i,j} [|I^{(m')}(i,j|x)|]]]
                    }
 \end{align}
 where $\Omega $ denotes the set of all samples. 
the relative interaction strength $J^{(m)} $
is computed over all pairs of input variables in all
samples. $J^{(m)} $ is normalized by the average value of interaction strength. The distribution of $J^{(m)} $ measures the distribution of the complexity of interactions encoded in DNNs. 

Given the context $S$, consider its subset $T \subseteq S$, which forms a coalition to represent a specific inference pattern $T\cup \{i, j\}$. Let $R^T (i,j)$ quantify the
marginal reward obtained from the inference pattern of an object. All interaction effects from smaller coalitions $T' \subsetneqq $ T are removed from $R^T (i,j)$. We analyze the mechanisms of several representative data augmentation methods based on these concepts.

         \begin{figure}[t]
          \includegraphics[width=0.48 \textwidth]{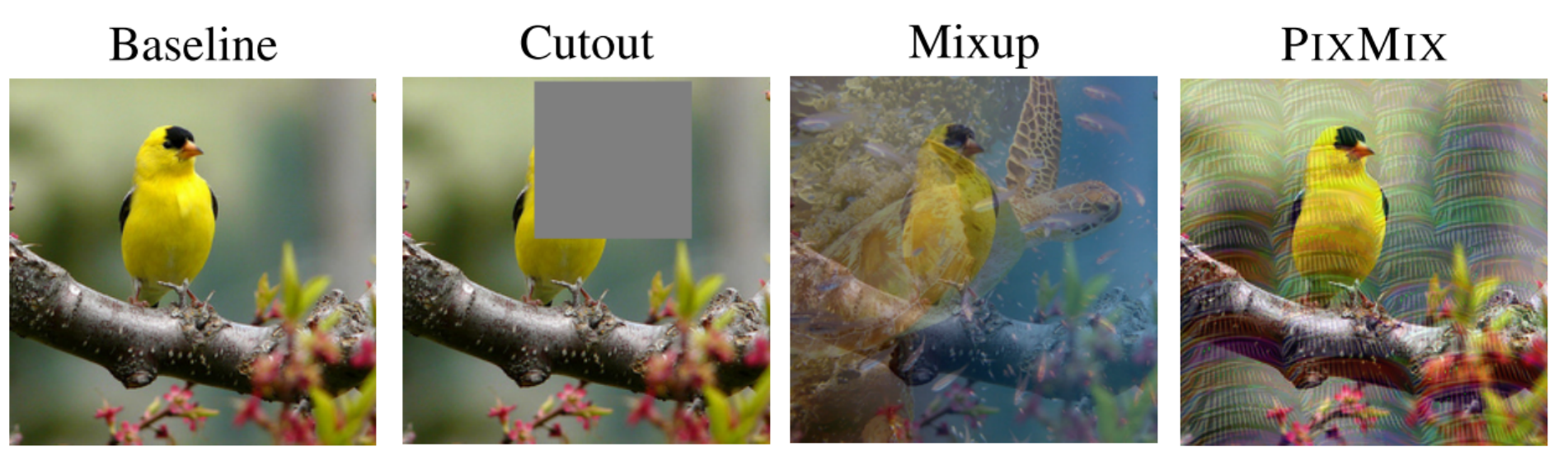}
                 \vspace{-8mm}   
          \caption{Visual representation of three popular data augmentation methods \cite{hendrycks2022pixmix}} \label{augfig1}
                 \vspace{-6mm}   
        \end{figure}  

\vspace{-4mm}
        \subsection{Flip/Crop}
        \vspace{-2mm}
Because almost all the existing deep neural networks have the convolutional modules, when the picture information is processed, there is a transitional infariance property. Therefore, small local visual concepts, such as bird pakes, will not affect the recognition of deep neural networks when the picture is flipped and cut. Viewing from the perspective of the game interaction, the flip and cutting operation does not change marginal award $R^T(i,j)$ obtained from the inference pattern the local set $T$:
\begin{align}
    R^T(i,j) \approx R^{T'}(i,j), s.t. \ |T| << |S|, |T'| << |S|
    \end{align}
where $T'$ is new subset $T' \subseteq S$, which forms a coalition to represent a specific inference pattern $T'\cup \{i, j\}$. If we cut the picture, the number of useful patterns in the picture will decrease. For example, the picture is a bird standing on the tree. If we cut off the tree and only retain the part of the bird, it will have the effect with the Cutout method.

However, when we are not concerned about the local object, but the global information, the impact of the flip is no longer small. After the picture is flipped, the information location of the context used to calculate the interaction of the game has changed. For example, the tree is on the right side of the bird in the original picture, and the tree is turned to the left side of the bird during a certain epoch of training. At this time, DNN learns a richer game interaction information, and this kind of information is often global instead of local. Therefore, the mid-order and high-order game interactions is encouraged.

\vspace{-4mm}
\subsection{Cutout}
\vspace{-2mm}
Cutout \cite{devries2017improved} is a random mask-based data augmentation method. This method randomly cuts some pixels, which can be considered a random dropout operation. Intuitively, cutting pixels will remove some reasoning modes in pictures. As shown in Figure \ref{augfig1}, the method randomly cuts off the head of the bird. Theoretically, \cite{zhang2020interpreting} proves when $ r = |S'| \leq k $:
\begin{align}
   \frac{I_{cutout}^{(k)}(i,j)}{I^{(k)}(i,j)} = \frac{\sum_{0\leq q\leq r} C_r^q \mathbb{E}_{T \subseteq N\backslash \{ i,j\}, |T|=q}R^{T}(i,j)}{\sum_{0\leq q\leq k} C_k^q \mathbb{E}_{T \subseteq N\backslash \{ i,j\}, |T|=q}R^{T}(i,j)} \leq 1
\end{align}
which shows that the number of inference patterns usually significantly decreases when we use dropout to remove $k-r, r\sim B(k, p)$, activation units. Empirically, experiments in \cite{zhang2020interpreting} show that low-order game interactions will be more compressed than mid- and high-order game interactions. Therefore, when we use relative interaction strength $J^{(m)}$,  the interaction of low-order will be compressed, $J^{(m)} $ of mid-order and high-order will be activated. Our experiments \ref{ex1} also show that when using Cutout as a data augmentation method, mid- and high-order game interaction accounts for more in total interactions.

\vspace{-4mm}
\subsection{Mixup}
\vspace{-2mm}

The Mixup \cite{zhang2018mixup} is a simple data augmentation strategy for computer vision data. Through linear transformation of input data as follows:
\begin{align}
    \hat X = \lambda \cdot X_0 + (1 - \lambda) \cdot X_1
\end{align}
This method can increase the generalization ability of the model, and can improve the robustness of the model to the adversarial attack. At the same time, the author find that the use of Mixup in various semi-supervised tasks can greatly improve the generalization ability of the model.

The input image can be intuitively understood as a mixture of two images, which usually does not have a significant impact on local details. Intuitively, when a picture of a turtle and a bird are mixed, the local information often does not change. We can still identify the bird through the local information, such as the bird's head. However, the context information used for interaction has changed significantly, and the underwater world scene appears in the bird picture when the background of plants is weakened. Therefore, during the Mixup operation, local context information is significantly affected, for example, plants that often appear with birds are masked. Therefore, the interaction generated by local context information decreases.  

Theoretically, it has been proven that $I^{(m)}(i,j)$ satisfies the following desirable linear property:
\begin{align}
   I_{w}^{(m)}(i,j) = I_{u}^{(m)}(i,j) + I_{v}^{(m)}(i,j)
\end{align}
where $I_{w}^{(m)}(i,j)$ is defined in \cite{zhang2020interpreting}. If two independent games $u$ and $v$ are combined, i.e., for $ \forall S \subseteq N, w(S) =
u(S) + v(S)$, then the multi-order interaction of the combined game equals to the sum of multi-order interactions derived from $u$ and $v$.
We can assume that when performing a Mixup operation, two different input variables are independent. In this way, we can mark the $s$ order game interaction of two input variables $X_0$, $X_1$ as $I_{1}^{(s)}(i,j)$ and $I_{2}^ {(s)}(i,j)$. According to the linear property, when two variables are mixed, the total game interaction can be written as $I_{1}^{(s)}(i,j) + I_{2}^{(s)}(i,j)$. Let $R^T (i,j)$ quantify the
marginal reward obtained from the inference pattern of an object. All interaction effects from smaller coalitions $T' \subsetneqq $ T are removed from $R^T (i,j)$. Then, we have Eq. \ref{eq:mixup}.
\vspace{-1mm}
\begin{align}
\label{eq:mixup}
  I_{i,j}^{(k)} &=  I_{1}^{(k)}(i,j) + I_{2}^{(k)}(i,j)\\
    &= \mathbb{E}_{S \subseteq N \backslash \{ i,j\},|S|=k} \Bigl[ \sum_{{T_1}\subseteq S } R^{T_1}(i,j) + \sum_{{T_2}\subseteq S } R^{T_2}(i,j) \Bigr] \  
    \\
    &= \frac{1}{C_{n-2}^{k}} \sum_{ \substack{S \subseteq N \backslash \{ i,j\} \\ |S|=k}} \Big[ \sum_{{T_1}\subseteq S } R^{T_1}(i,j)  + \sum_{{T_2}\subseteq S } R^{T_2}(i,j) \Big] \\
    &= \frac{1}{C_{n-2}^{k}} \sum_{0 \leq q \leq k} \Big[ \sum_{ \substack{S \subseteq N \backslash \{ i,j\} \\ |S|=k}} \Big(\sum_{ \substack{ {T_1}\subseteq S \\ |T_1|=q }} R^{T_1}(i,j) \\ &+ \sum_{ \substack{ {T_2}\subseteq S \\ |T_2|=q }} R^{T_2}(i,j) \Big) \Big] \\
    &= \frac{1}{C_{n-2}^{k}} \sum_{0 \leq q \leq k} \Big[ \sum_{ \substack{T_1 \subseteq N \backslash \{ i,j\} \\ |T_1|=q}} \sum_{ \substack{ {T'_1}\subseteq N \backslash \{ i,j\} \backslash T_1 \\ |T'_1|=k-q }} R^{T_1}(i,j)  \\ &+  \sum_{ \substack{T_2 \subseteq N \backslash \{ i,j\} \\ |T_2|=q}} \sum_{ \substack{ {T'_2}\subseteq N \backslash \{ i,j\} \backslash T_2 \\ |T'_2|=k-q }} R^{T_2}(i,j)) \Big] \\
    &=\frac{1}{C_{n-2}^{k}} \sum_{0 \leq q \leq k} \Big[ \sum_{ \substack{T_1 \subseteq N \backslash \{ i,j\} \\ |T_1|=q}} C_{n-2-q}^{k-q} R^{T_1}(i,j)  \\ &+ \sum_{ \substack{T_2 \subseteq N \backslash \{ i,j\} \\ |T_2|=q}} C_{n-2-q}^{k-q} R^{T_2}(i,j) \Big] \\
    &= \frac{1}{C_{n-2}^{k}} \sum_{0 \leq q \leq k} \Big[ C_{n-2}^{q}  C_{n-2-q}^{k-q} \mathbb{E}_{\substack{T_1 \subseteq N \backslash \{ i,j\} \\ |T_1|=q}} R^{T_1}(i,j) \\ &+ C_{n-2}^{q}  C_{n-2-q}^{k-q} \mathbb{E}_{\substack{T_2 \subseteq N \backslash \{ i,j\} \\ |T_2|=q}} R^{T_2}(i,j) \Big] \\
    &= \sum_{0 \leq q \leq k} C_k^q  \Big[ \mathbb{E}_{\substack{T_1 \subseteq N \backslash \{ i,j\} \\ |T_1|=q}} R^{T_1}(i,j) + \mathbb{E}_{\substack{T_2 \subseteq N \backslash \{ i,j\} \\ |T_2|=q}} R^{T_2}(i,j) \Big] 
\vspace{-4mm}
\end{align}
where $S=T_1 \cup T'_1, T_1 \cap T'_1=\emptyset$ and $S=T_2 \cup T'_2, T_2 \cap T'_2=\emptyset$. Then, we only need to focus on the inference pattern $T_1$ and $T_2$ in the two images. From the Figure \ref{augfig1}, we find that local objects such as the beak and the turtle's head are not affected. But the textures of the details are overridden, for example the texture of the bird feathers changes. Therefore, we can reasonably assume that when $q$ is much smaller than $k$, it is affected by more detailed textures, because low-order interactions often represent common features and are shared by different categories \cite{cheng2021game}. Because the detail texture related to the target class is overridden when $q$ is much smaller than $k$, the marginal reward obtained from the inference pattern will decrease:
\begin{align}
 \mathbb{E}_{\substack{T_1 \subseteq N \backslash \{ i,j\} \\ |T_1|=q}} R^{T_1}(i,j) +  \mathbb{E}_{\substack{T_2 \subseteq N \backslash \{ i,j\} \\ |T_2|=q}} R^{T_2}(i,j)
  < \\  \mathbb{E}_{\substack{T \subseteq N \backslash \{ i,j\} \\ |T|=q}} R^{T}(i,j), s.t. \ q << k.
\end{align}
However, when $q$ is gradually increased, the influence of medium-sized picture objects becomes significant. Because the Mixup operation does not cover these objects, we can  assume that $T_1$ and $T$ contain similar inference patterns at this time, then:
\begin{align}
 \mathbb{E}_{\substack{T_1 \subseteq N \backslash \{ i,j\} \\ |T_1|=q}} R^{T_1}(i,j) 
  	\approx \mathbb{E}_{\substack{T \subseteq N \backslash \{ i,j\} \\ |T|=q}} R^{T}(i,j)
   \end{align}
   Then the marginal reward obtained from the inference pattern $T_2$ leads to an increase in game interaction.

             \begin{figure}[t]
          \includegraphics[width=0.5 \textwidth]{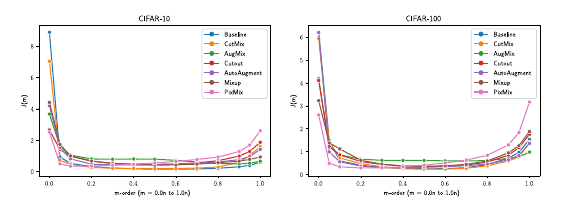}
                 \vspace{-8mm}   
          \caption{
          The relative game interaction strengths of Wide ResNet trained on CIFAR-10/100 using different data augmentation methods.} \label{fig1}
              \vspace{-4mm}   
        \end{figure} 
        
\vspace{-4mm}
\subsection{PixMix}
\vspace{-2mm}
When using PixMix, as shown in the figure \ref{app_a}, the picture is artificially added with more complex detail textures, and more inference patterns are added to the picture. The purpose of this data augmentation is to increase the complexity of the picture. Looking at the image, we can reasonably assume that the same region contains more complex detail textures as inference patterns. Such complex details have a greater impact when there is less contextual information. For example, when observing the eyes in a face picture alone, if complex textures are added to the eyes, the deep neural network will learn not only the single object of the eyes, but also the complex textures. However, if there is a lot of context information, that is, the order of game interaction is high, local complex textures will not have an impact. Similar to the action mechanism of Mixup, the local texture details added by the Pixmix method are often irrelevant to the target class. This kind of unrelated complex detail texture makes DNN tend to use low-order interaction with lack of context information when learning, but uses global interaction information with richer context information. We can write this mechanism as:
\begin{align}
 \mathbb{E}_{\substack{T' \subseteq N \backslash \{ i,j\} \\ |T'|=q}} R^{T'}(i,j) 
  	< \mathbb{E}_{\substack{T \subseteq N \backslash \{ i,j\} \\ |T|=q}} R^{T}(i,j), s.t. |T| << |S|
\end{align}
where $T'$ is the new subset $T' \subseteq S$, which forms a coalition to represent a specific inference pattern $T'\cup \{i, j\}$ when Pixmix is used.  As a result, low-order game interaction is suppressed. When calculating the relative game interaction $J^(m)$, mid- and high-order game interactions are encouraged.
\vspace{-4mm}
\subsection{A Proxy for Safety and Robustness Metrics}
\vspace{-2mm}
Based on the mining of the relationship between the safety and robustness measures and the relative interaction strength $J^{(m)}$ defined in \cite{deng2021discovering}, we propose a proxy called \textbf{adjusted mid-order relative interaction strength (AMRIS)} for the effectiveness of data augmentation on model robustness improvement, defined as follows:
\begin{equation}
\label{eq:proxy}
    AMRIS(a,b,c) = \sqrt {
    \frac{1}{\max {\mathbf{J}} - \min {\mathbf{J}}}
    \frac{\sum_{m=\lfloor bn \rfloor}^{\lfloor cn \rfloor} J^{(m)}}
    {\sum_{m=0}^{\lfloor an \rfloor} J^{(m)}}
    }
\end{equation} 
where $0 \leq a \leq b \leq c \leq 1$, and $a$,$b$,$c$ are parameters that control how the order of interactions is selected.
$J^{(m)}$ is the relative interaction strength, and $n$ is the total number of variables. $\mathbf{J} = (J^{(1)},...,J^{(m)})$ is relative game interactions of all orders. As shown in Figure \ref{fig1}, \textbf{the data augmentation methods that bring a more robust model tend to have lower low-order interactions and higher mid-order interactions.} But \cite{ren2021unified} demonstrate that adversarial attacks mainly affect high-order interactions. \textbf{In fact, $AMRIS$ measures the strength of mid-order interactions relative to lower-order interactions.} The method to choose $a, b, c$ is grid search, and the evaluation metric is the mean correlation between $AMRIS$ and other robustness metrics.

\vspace{-2mm}
\section{Experiments}
\vspace{-2mm}
\subsection{Basic Settings}
\vspace{-2mm}
Our experiments are mainly coded in PyTorch 1.11.0 and run on Nvidia's GPUs and Linux 5.4.0. The experiments of CIFAR-10/100 mainly run on the A5000 GPU with 24GB memory and the A40 GPU with 48GB memory. The training datasets we use include CIFAR-10 and CIFAR-100 \cite{krizhevsky2009learning}. To evaluate the safety and robustness of DNNs, we also use the CIFAR-10/100-C, CIFAR-10/100-$\rm \bar C$ \cite{hendrycks2018benchmarking, mintun2021interaction}.
For a fair comparison, the models we use include vanilla 40-4 Wide ResNet \cite{WideResNet}. The drop rate of the 40-4 Wide ResNet we use is 0.3 and the initial learning rate is 0.05 which follows the cosine schedule \cite{loshchilov2016sgdr}.  For the parameters of PixMix, we follow the settings in the original paper \cite{hendrycks2022pixmix}. 

\vspace{-4mm}   
\subsection{Relative Game Interaction Strength}
\vspace{-2mm}   
\label{ex1}
         \begin{figure}[t]
          \includegraphics[width=0.45\textwidth]{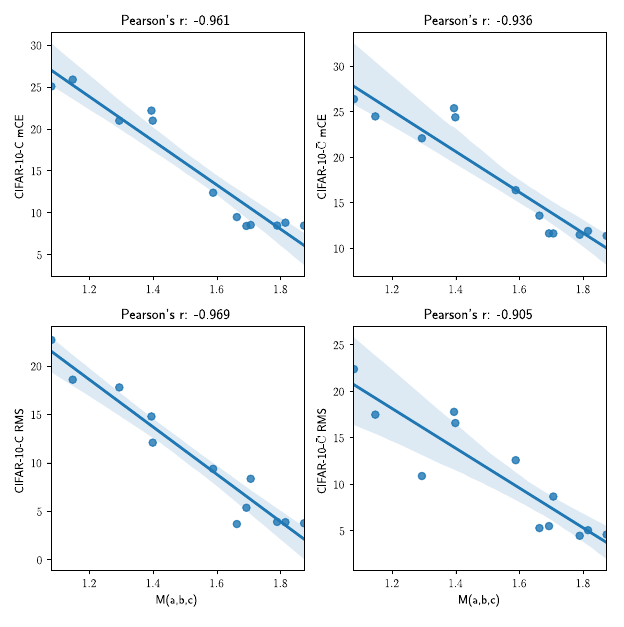}
                 \vspace{-6mm}   
          \caption{Relationship between the proxy $AMRIS$ and some safety and robustness metrics when using CIFAR-10-C and CIFAR-10-$\rm \bar C$.} \label{fig3}
                        \vspace{-6mm}   
        \end{figure}      

We explore some models that train with different data augmentation technologies provided by \cite{hendrycks2019augmix,hendrycks2022pixmix,cubuk2019autoaugment,TrivialAugment,yun2019cutmix,devries2017improved}. There are six data augmentation techniques: Cutout, CutMix, AutoAugment, Mixup, AugMix, and PixMix. Models trained based on these techniques outperform standard trained baseline models on safety and robustness metrics. We calculate the relative game interaction strength of these models and show the results of these models in Figure \ref{fig1}. Compared with the baseline model, we find that the common feature of models trained with data augmentation is to weaken low-order relative game interaction strength and encourage mid-order and high-order relative game interaction strength. This phenomenon suggests that low-order interactions contribute less to the model's robustness than mid-order and high-order interactions. Our experiments also validate the conclusion in \cite{deng2021discovering} that mid-order interactions are hard to learn for DNNs without additional constraints. The best method for balancing various safety and robustness metrics is PixMix, and it has the smallest low-order interactions and the highest high-order interactions. This inspires us that if we want to improve the robustness of a deep learning model, we need to \textbf{pay more attention to the mid-order and high-order interactions}, and have \textbf{a global understanding like human cognition rather than limited to small local information}.

          \begin{figure}[t]
          \centering
          \includegraphics[width=0.5 \textwidth]{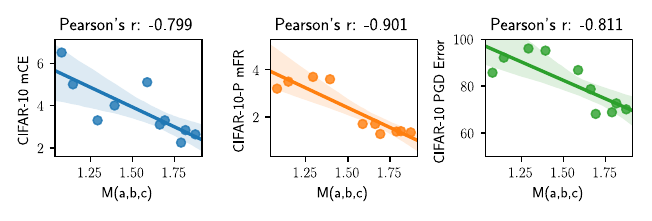}
                 \vspace{-8mm}   
          \caption{
          Relationship between the proxy $AMRIS$ and some other safety and robustness metrics when using CIFAR-10.} \label{appfig1}
                        \vspace{-4mm}   
        \end{figure}
        
        \begin{figure}[t]
        \centering
          \includegraphics[width=0.5 \textwidth]{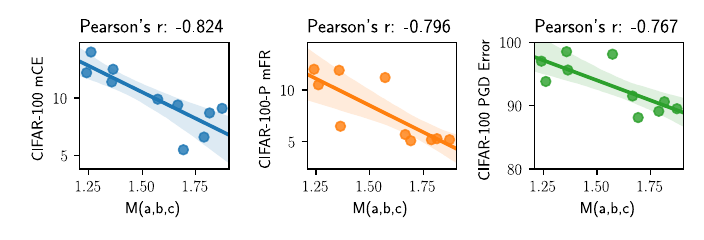}
                        \vspace{-8mm}   
          \caption{Relationship between the proxy $AMRIS$ and some other safety and robustness metrics when using CIFAR-100.} 
          \label{appfig2}
                        \vspace{-6mm}   
        \end{figure}

          \vspace{-4mm}   
\subsection{The Proxy for Safety and Robustness}
  \vspace{-2mm}   
The relationship between our proxy and some safety and robustness metrics is shown in Figure \ref{fig3}, \ref{appfig1}, and \ref{appfig2}. The mCE means mean corruption error, PGD means using projected gradient descent (PGD) \cite{madry2017towards} to generate untargeted perturbations. The RMS means calibration error \cite{nguyen2015posterior} which is defined as $\sqrt{\mathbb{E}_C [(\mathbb{P}(Y = \hat Y |C =c) - c)^2]}$, where $C$ is the confidence to predict $\hat Y$ correctly. 

 The Pearson correlation coefficient values of our $AMRIS(a, b, c)$ and various metrics are almost all above 0.9 in CIFAR-10, only a few values are between 0.8 and 0.9, indicating that the linear correlation is significant. There is a significant linear correlation between the proxy we propose and a variety of popular safety and robustness measures within a certain range. \textbf{Our proxy $AMRIS(a, b, c)$ can be used to estimate the robustness without calculating many complex metrics of a model}, it's an efficient global estimation tool for follow-up research on safe and robust AI. When using our proposed agent, only the relative game interaction strength needs to be calculated without using additional datasets. \textbf{Our proxy avoids the complexity of using various additional datasets and their biases}.
\vspace{-4mm}
 \section{Conclusion}
\vspace{-2mm}

This study thoroughly explored the impact of data augmentation techniques on the safety and robustness of DNNs using the CIFAR-10 and CIFAR-100 datasets. The central observation was the shift from low- and high-order to a focus on mid-order relative game interaction strengths, revealing its importance in model robustness.

Furthermore, the introduction of the $AMRIS$ proxy offers a streamlined approach to evaluate model robustness without complex metric computations, reducing the reliance on additional datasets. This research underscores the significance of a holistic understanding of data augmentation techniques and provides essential insights for future AI research on safety and robustness.
\vfill\pagebreak

\bibliographystyle{IEEEbib}
\bibliography{strings,refs}

\end{document}